\documentclass[10pt,twocolumn,letterpaper]{article}

\usepackage{cvpr}
\usepackage{times}
\usepackage{epsfig}
\usepackage{graphicx}
\usepackage{amsmath}
\usepackage{amssymb}
\usepackage{multirow}
\usepackage{cite}
\usepackage{subcaption}

\usepackage[pagebackref=true,breaklinks=true,letterpaper=true,colorlinks,bookmarks=false]{hyperref}

\usepackage{amsthm}

\theoremstyle{definition}
\newtheorem{definition}{Definition}

\cvprfinalcopy 

\def\cvprPaperID{1657} 

\ifcvprfinal\fi
\begin{document}
	
	\title{Inertial-aided Rolling Shutter Relative Pose Estimation}
	
	\author{Chang-Ryeol Lee\and 
		Kuk-Jin Yoon\and
		Gwangju Institute of Science and Technology (GIST), South Korea\\
		{\tt\small \{crlee, kjyoon\}@gist.ac.kr}
	} 
	
	\maketitle
	
	\begin{abstract}
		Relative pose estimation is a fundamental problem in computer vision and it has been studied for conventional global shutter cameras for decades.  
		However, recently, a rolling shutter camera has been widely used due to its low cost imaging capability and, since the rolling shutter camera captures the image line-by-line, the relative pose estimation of a rolling shutter camera is more difficult than that of a global shutter camera. 
		In this paper, we propose to exploit inertial measurements (gravity and angular velocity) for the rolling shutter relative pose estimation problem.
		The inertial measurements provide information about the partial relative rotation between two views (cameras) and the instantaneous motion that causes the rolling shutter distortion. 
		Based on this information, we simplify the rolling shutter relative pose estimation problem and propose effective methods to solve it.
		Unlike the previous methods, which require 44 (linear) or 17 (nonlinear) points with the uniform rolling shutter camera model, the proposed methods require at most 9 or 11 points to estimate the relative pose between the rolling shutter cameras.
		Experimental results on synthetic data and the public PennCOSYVIO dataset show that the proposed methods outperform the existing methods.
	\end{abstract}

	\section{Introduction}\label{sec:intro}

	Rolling shutter cameras, which capture images line-by-line, have become popular due to their low-cost imaging capabilities.
	However, the rolling shutter cameras cause undesirable artifacts if they move during the image capture or if they capture the dynamic scenes or objects, due to the line-by-line imaging characteristics.
	The rolling shutter distortion has a critical influence on geometric vision applications such as structure-from-motion (SfM), simultaneous localization and mapping (SLAM), and dense 3D reconstruction~\cite{Albl:ECCV:2016}.
	Resolving the rolling shutter distortion has been studied in the computer vision and robotics community over the past few years~\cite{Hedborg:ICCVW:2011,Hedborg:CVPR:2012,Saurer:ICCV:2013,Guo:RSS:2014,Kim:ICRA:2016}.
	Most of previous works focused on video applications based on temporal interpolation of camera poses.
	However, relative pose and SfM problems using unordered still images captured by rolling shutter cameras have only recently begun to study.
	Dai \etal~\cite{Dai:CVPR:2016} defined the rolling shutter relative pose problem,
	and they proposed linear and nonlinear algorithms with linear and uniform rolling shutter camera models for estimating the relative pose of rolling shutter cameras.
	Albl \etal proposed an absolute pose estimation algorithm of rolling shutter cameras~\cite{Albl:CVPR:2015}, and analyzed the degeneracy of the rolling shutter structure from motion and proposed a scheme to avoid this situation~\cite{Albl:ECCV:2016}.

	%
	In this paper, we tackle the rolling shutter relative pose estimation problem defined in~\cite{Dai:CVPR:2016} and take one step further by the aid of inertial measurements.
	Dai \etal proposed the linear algorithm that requires at least 44 points and the nonlinear algorithm that requires 17 points with the uniform rolling shutter camera model.
	However, random sample consensus (RANSAC) with 44 points for relative pose estimation is prone to be sensitive to outliers, and it is time-consuming since they have to examine a lot of hypotheses.
	In addition, the nonlinear algorithm, which requires at least 17 points, is also time-consuming because it performs nonlinear least square optimization on 18 unknown variables for each iteration of RANSAC.
	Furthermore, estimating 18 unknown variables with just two images fall into the local minima.
	Therefore, it is necessary to lower the degree-of-freedom (DOF) of the problem for practical use of the rolling shutter relative pose estimation.
	In this sense, inertial measurements can be used to lower the DOF of the problem as in \cite{Fraundorfer:ECCV:2010, Lee:CVPR:2014, Albl:CVPR:2016}. 
	
	%
	In this paper, we propose a novel method to estimate the relative pose of rolling shutter cameras with the help of 6-DOF inertial measurements consisting of 3D gravity and 3D angular velocity measurements.
	The directions of gravity in two cameras can be converted to 2D relative rotation between two cameras.
	The angular velocity measurements directly provide instantaneous motion information that causes rolling shutter distortion.
	We derive five algorithms based on angular, linear, and uniform rolling shutter camera models, using only angular velocity measurements or using both angular velocity and gravity measurements: 1) linear 9-point, 2) angular 5-point, 3) angular 3-point, 4) uniform 11-point, and 5) uniform 9-point algorithms.
	Finally, we refine the estimates using an alternative directing scheme that optimizes the decoupled variables alternately and iteratively.
	
	Our contributions can be summarized as follows.
	\begin{itemize}	
		\vspace{-2.5mm} 
		\setlength\itemsep{-1mm}
		\item Introduction of the use of inertial measurements for rolling shutter relative pose estimation problem
		\item Derivation of the five algorithms to use only angular velocity measurements and to use both gravity and angular velocity measurements.
		\item Refinement of rotation and translation estimates with an alternative directing scheme.
	\end{itemize}

	\section{Related Works}\label{sec:related_works}

	The rolling shutter camera was addressed in the Perspective-n-Point (PnP) problem to estimate the camera pose with given 2D projections corresponding to 3D point clouds.
	Ait-Aider \etal~\cite{Ait:ECCV:2006} proposed to estimate the pose and speed of fast-moving objects in a single image with a given 2D-3D matching.
	They proposed nonlinear and linear models for non-planar and planar objects.
	Magerand \etal~\cite{Magerand:ECCV:2012} extended this study by suggesting a polynomial uniform rolling shutter camera model and solving the problem through constrained global optimization.
	Albl \etal~\cite{Albl:CVPR:2015} proposed a method to estimate the camera pose with only 6 points based on the double linearized rolling shutter camera model.

	Besides, SfM and SLAM based on a monocular rolling shutter camera have been studied.
	Klein and Murray~\cite{Klein:ISMAR:2009} used the constant velocity model in the SLAM framework to predict and correct rolling shutter distortion occurring in the next frame.
	Hedborg \etal~\cite{Hedborg:ICCVW:2011,Hedborg:CVPR:2012} applied the rolling shutter camera projection model to the bundle adjustment of SfM.
	Their key idea is to exploit temporal continuity of the camera motion on video input to deal with rolling shutter distortion.
	Saurer \etal~\cite{Saurer:ICCV:2013} dealt with the rolling shutter distortion in dense 3D reconstruction.
	Albl \etal~\cite{Albl:ECCV:2016} analyzed the degeneracy of the rolling shutter SfM and suggested how to avoid the degeneracy when shooting videos.
	Recently, Ito and Okatani~\cite{Ito:CVPR:2017} derived the degeneracy of rolling shutter SfM as a general expression through a self-calibration-based approach.
	Zhuang \etal~\cite{Zhuang:ICCV:2017} proposed a constant acceleration model for relative pose estimation and image rectification in two consecutive images.

	On the other hand, the methods to utilize an inertial measurement unit (IMU) to deal with rolling shutter distortion in visual odometry (VO) and SLAM have been also studied.
	Jia and Evans~\cite{Jia:MMSP:2012} proposed a method to estimate the camera orientation using gyroscope measurements and to correct the rolling shutter distortion of the image.
	Guo~\etal~\cite{Guo:RSS:2014} applied a rolling shutter camera projection model to a visual-inertial odometry framework that uses IMUs and cameras to estimate egomotion.
	They estimated the readout time of the rolling shutter camera as well as the time delay between the IMU and the camera.
	Albl~\etal~\cite{Albl:CVPR:2016} proposed a method to improve the speed and accuracy of the method in \cite{Albl:CVPR:2015} using the gravity obtained from the IMU.

	In addition, IMUs are often used to solve conventional relative or absolute pose estimation problems for global shutter cameras.
	There have been studies to estimate relative pose with partially known orientation angle between two cameras \cite{Fraundorfer:ECCV:2010,Li:IROS:2013}, or with known vertical direction \cite{Kukelova:ACCV:2011,Lee:CVPR:2014}.

	The relative pose estimation is a fundamental problem and of eminent importance in SfM.
	To the best of our knowledge, this is first work to exploit an IMU for the relative pose estimation of the rolling shutter cameras.
	
	\section{Problem Formulation} \label{sec:problem}

\begin{figure}[tb]
		\centering	
		\includegraphics[width=0.99\linewidth]{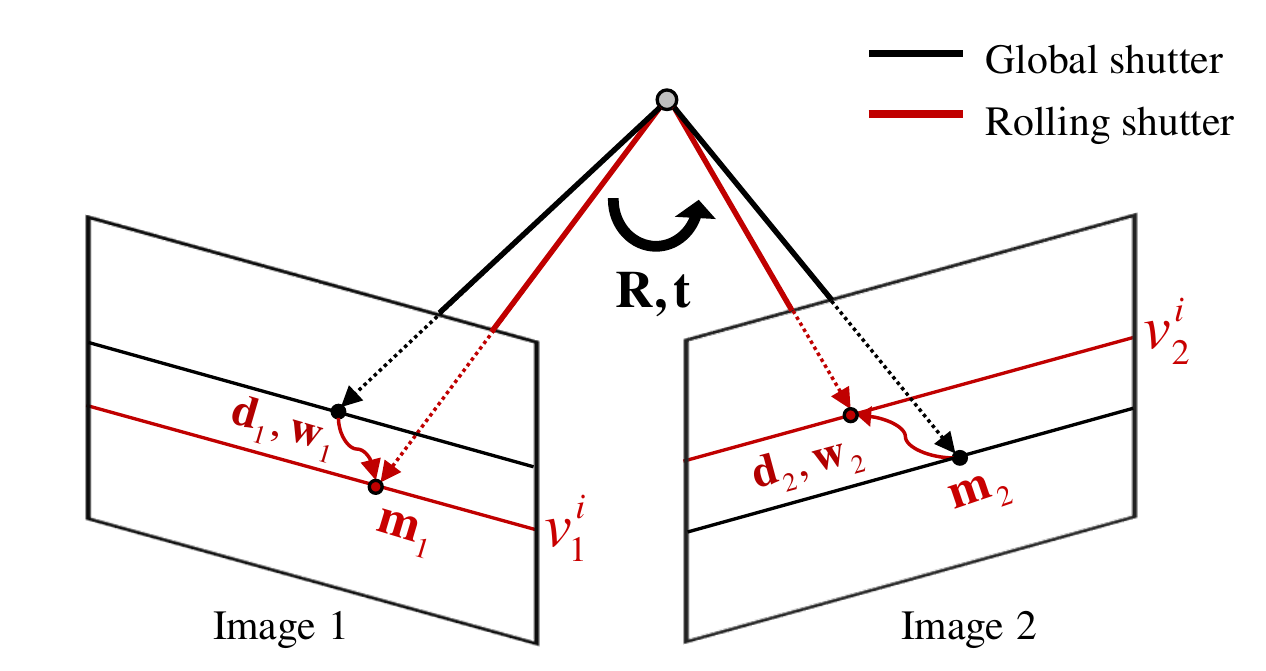}	
		\caption{Two-view geometry of a rolling shutter camera. }		
		\label{fig:rs_two_view_geometry} 
	\end{figure}

	In this paper, we assume that a rolling shutter camera acquires images in a row-by-row manner, not column-by-column, and the motion from the first row to the last row of the rolling shutter image follows the constant velocity model.
	In this setting, the rotational and translational motion in each row, which causes the rolling shutter distortion, is represented by angular and linear velocities $\mathbf{w} \in \mathbb{R}^{3},\mathbf{d} \in \mathbb{R}^{3}$, a rolling shutter readout time $\lambda_r$, and the index of the row $v$.
	Figure~\ref{fig:rs_two_view_geometry} shows the geometry between two rolling shutter images in comparison with global shutter camera geometry.
	
	The rolling shutter epipolar constraint between two image points is described by the rolling shutter essential matrix:
	\begin{equation}
	\mathbf{m}_{2}^{\top} \mathbf{E}_{r} \mathbf{m}_{1} = 0 ,
	\label{eq:rolling_shutter_epipolar_geometry}
	\end{equation}
	where $\mathbf{m}_{1},\mathbf{m}_{2} \in \mathbb{P}^2$ are corresponding points in the normalized camera coordinate, and  $\mathbf{E}_{r}$ is the rolling shutter essential matrix.
	Three camera models have been commonly used to explain the rolling shutter camera geometry: 1) angular, 2) linear, and 3) uniform models.
	Here, we introduce the rolling shutter relative pose problem formulations with the three models.

	At first, the angular rolling shutter camera model assumes a linear velocity to be zero.
	Thus, the rolling shutter essential matrix $ \mathbf{E}_r $ considers only $\textit{angular}$ velocities in the two images as 
	\begin{equation}
	\mathbf{E}_{r} = \left(\mathbf{I} + v_2 \lambda_{r} \lfloor \mathbf{w}_2 \rfloor_{\times} \right)^{\top}  \mathbf{R} \left(\mathbf{I} + v_1 \lambda_{r} \lfloor \mathbf{w}_1 \rfloor_{\times} \right) \lfloor \mathbf{t} \rfloor_{\times} ,
	\end{equation}
	where $\mathbf{R} \in SO(3)$ is the relative rotation, $\mathbf{t} \in \mathbb{R}^{3}$ is the relative translation.
	%
	%
	%
	$\lfloor \cdot \rfloor_{\times}$ is the matrix form of the cross product.

	Secondly, the linear model assumes an angular velocity to be zero.
	Thus, unlike the angular model, the rolling shutter essential matrix considers only $\textit{linear}$ velocities in two images as 
	\begin{equation}
	\mathbf{E}_{r} = \mathbf{R} \lfloor \mathbf{t} - v_1 \lambda_{r} \mathbf{d}_{1}  + v_2 \lambda_{r} \mathbf{d}_{2}  \rfloor_{\times}  ,
	\label{eq:rolling_shutter_essential_matrix}
	\end{equation}
	where $\mathbf{d}_i$ is the linear velocity at which each image is captured.

	Finally, the most popular uniform model considers both $\textit{angular}$ and $\textit{linear}$ velocities.
	With this model, the rolling shutter essential matrix is defined as:
	\begin{equation}
	\begin{split}
	\mathbf{E}_{r} & = \mathbf{R}_{r} \lfloor \mathbf{t} - v_1 \lambda_{r} \mathbf{d}_{1}  + v_2 \lambda_{r} \mathbf{d}_{2}  \rfloor_{\times} ,  \\
	\mathbf{R}_{r} & = \left(\mathbf{I} + v_2 \lambda_{r} \lfloor \mathbf{w}_2 \rfloor_{\times} \right)^{\top}  \mathbf{R} \left(\mathbf{I} + v_1 \lambda_{r} \lfloor \mathbf{w}_1 \rfloor_{\times} \right) .
	\end{split}
	\label{eq:rolling_shutter_essential_matrix}
	\end{equation}
	With these models, we exploit the inertial measurements to estimate the relative pose of the rolling shutter camera.
	Therefore, our rolling shutter relative pose estimation problem is defined as follows:
	\begin{definition}{}
		Given a calibrated rolling shutter camera--IMU system (i.e., known intrinsics (of a camera), extrinsics (between a camera and an IMU), and rolling shutter readout times), inertial measurements (gravity and angular velocity from the IMU) and image point correspondences between the two rolling shutter camera--IMU system frames, find the relative pose between the two rolling shutter camera--IMU frames (i.e., $\mathbf{R}, \mathbf{t}$).
	\end{definition}

	\section{Proposed Methods}

	We exploit inertial measurements for three different rolling shutter relative pose estimation problems, depending on the given model: 1) angular, 2) linear, and 3) uniform models.
	The inertial measurements consist of gravity and angular velocity measurements $\mathbf{g}$ and $\mathbf{w}$.
	The number of minimal points for rolling shutter relative pose estimation is determined by how the two kinds of inertial measurements are used for given three models.
	Consequently, we present five different algorithms that can be chosen appropriately for applications.

	We first explain how to use the gravity and angular velocity measurements for the rolling shutter relative pose estimation problems (Sec~\ref{subsec:gravity}),
	and we describe the derivation of the five algorithms (Sec~\ref{subsec:minimal_points}).
	In the section, we explain the linear rolling shutter relative pose problem using gravity measurements.
	This algorithm requires 9 points to estimate relative velocities and linear velocities in two frames.
	Then, we explain how to use only angular velocity measurements for an angular rolling shutter relative pose problem, and how to use both angular velocity and gravity measurements.
	The use of both angular velocity and gravity measurements reduces the complexity of the problem than using only the angular velocity measurements. 
	Using only the angular velocity measurements to solve the problem requires 5 points, and using both measurements requires 3 points (\ie 3 correspondences).
	Similarly, for the uniform rolling shutter relative pose problem, we describe the 11-point algorithm, which uses only angular velocity measurement, and the 9-point algorithm, which uses both measurements.
	Table~\ref{table:proposed_method_summary} summarizes the rolling shutter camera models, the inertial measurements used, and the DOF of the problem. 
	In Sec~\ref{subsec:solver}, we explain how to estimate the solution of these five algorithms.
	%
	%
	Finally, we describe the refinement to reduce errors of estimates in Sec~\ref{subsec:refinement}.

	\begin{table}[t]
		\setlength{\tabcolsep}{2pt}
		\footnotesize
		\caption{Summary of the proposed algorithms} 
		\label{table:proposed_method_summary}
		\begin{tabular}{|c|c|c|c|c|}
			\hline
			\begin{tabular}[c]{@{}c@{}} Rolling shutter \\ camera model \end{tabular} & \begin{tabular}[c]{@{}c@{}}Model\\   parameters\end{tabular} & \begin{tabular}[c]{@{}c@{}}Given \\ IMU \end{tabular} & Degree of freedom          & \begin{tabular}[c]{@{}c@{}}Minimal\\ points\end{tabular} \\ \hline
			Linear  & $\mathbf{R}, \mathbf{t}, \mathbf{d}_1, \mathbf{d}_2$  & $\mathbf{g}$   & 3+3+3+3 -2 = 10  & 9 \\ \hline
			\multirow{2}{*}{Angular} & \multirow{2}{*}{ $\mathbf{R}, \mathbf{t}, \mathbf{w}_1, \mathbf{w}_2$}   & $\mathbf{w}$ & 3+3+3+3 -6 = 6 & 5 \\ \cline{3-5}
			& & $\mathbf{g,w}$ & 3+3+3+3 -6-2 = 4 & 3  \\ \hline
			\multirow{2}{*}{Uniform} & \multirow{2}{*}{ \begin{tabular}[c]{@{}c@{}}  $\mathbf{R}, \mathbf{t}, \mathbf{d}_1, \mathbf{d}_2,$\\ $\mathbf{w}_1, \mathbf{w}_2$\end{tabular}  }  &  $\mathbf{w}$ & 3+3+3+3+3+3 -6 =12 & 11  \\ \cline{3-5}
			& & $\mathbf{g,w}$ & 3+3+3+3+3+3 -6-2 = 10 & 9 \\ \hline
		\end{tabular}		
		\vspace{1mm}
	\end{table}

	\subsection{Gravity and Angular Velocity Measurements} \label{subsec:gravity}

	Since the direction of the gravity measurements obtained from the IMU is a vertical direction in the world coordinate, it represents the slope (i.e., roll, pitch) of the IMU.
	The roll and pitch angles can be transformed to the camera coordinate through extrinsics between the IMU and the camera.
	Therefore, with the angles, the relative rotation between the two cameras can be expressed as:
	{
		\small
		\begin{equation}
		\begin{split}
		&\mathbf{R}  = 
		\mathbf{R}(\phi_2,\theta_2)^{\top} 
		\mathbf{R}(\psi)
		\mathbf{R}(\phi_1,\theta_1),
		\\
		& \mathbf{R}(\psi)
		=
		\begin{bmatrix}
		\cos\left(\psi\right) & 0 & -\sin\left(\psi\right) \\
		0 & 1 & 0 \\  
		\sin\left(\psi\right) & 0 & \cos\left(\psi\right)  
		\end{bmatrix},
		\\
		& \mathbf{R}(\phi,\theta) = \\
		&
		\begin{bmatrix}
		1 & 0 & 0 \\  
		0 & \cos\left(\phi\right) & -\sin\left(\phi\right) \\
		0 & \sin\left(\phi\right) & \cos\left(\phi\right)  
		\end{bmatrix}
		\begin{bmatrix}
		\cos\left(\theta\right) &  -\sin\left(\theta\right) & 0\\ 
		\sin\left(\theta\right) &  \cos\left(\theta\right)  & 0\\
		0 & 0 & 1 \\ 
		\end{bmatrix}
		,
		\end{split}
		\end{equation}
	}
	where $\mathbf{R}(\phi,\theta)$ is the 2D relative rotation obtained from the vertical direction, and $\mathbf{R}(\psi)$ is the rotation along the vertical direction.
	The angular velocity measurements obtained from the IMU are directly applied to angular and uniform rolling shutter relative pose estimation after being transformed to camera angular velocity through extrinsic parameters.

	\subsection{Minimal Algorithms} \label{subsec:minimal_points}

	To solve the rolling shutter relative pose estimation problem, we construct a homogeneous equation $\mathbf{A}\mathbf{x}=\mathbf{0}$ satisfying the rolling shutter epipolar constraint.
	The following section describes $\mathbf{A}$ and $\mathbf{x}$ which are organized by each algorithm.
	
	\vspace{3mm}
	\noindent $\textbf{Linear 9-point algorithm.}$
	Since the linear rolling shutter camera model does not take angular velocity into consideration, we do not use the angular velocity measurement of the IMU here, but only the gravity measurement.
	Given the gravities in two frames, the rolling shutter epipolar constraint can be written as:
	\begin{equation}
	{\mathbf{m}_{2}^{i}}^{\top} \mathbf{E}_{r} (\psi,\mathbf{t},\mathbf{d}_1,\mathbf{d}_2 ; \mathbf{g}_1,\mathbf{g}_2) \mathbf{m}_{1}^{i} = 0 .
	\label{eq:rolling_shutter_epipolar_geometry}
	\end{equation}
	The rolling shutter essential matrix consists of 10 unknown variables (3+3+3+3-2=10).
	However, since translation and linear velocity are up to scale, the number of minimal points for this problem becomes 9.
	We reformulate this equation as:
	\begin{equation}
	\begin{split}
	& p_{1}^i(\psi) t_x + p_{2}^i(\psi) t_y + p_{3}^i(\psi) t_z + p_{4}^i(\psi) d_{1x} \\ 
	& + p_{5}^i(\psi) d_{1y} + p_{6}^i(\psi) d_{1z} + p_{7}^i(\psi) d_{2x} \\
	& + p_{8}^i(\psi) d_{2y} + p_{9}^i(\psi) d_{2z} =   0 
	\end{split} ,	
	\end{equation}
	where superscript $i = \{1,2,3,\cdots,9\}$ is the index of a corresponding point,
	$p_{*}^i$ is a polynomial including $\psi$.
	We stack the equation as a linear matrix equation that decouples relative rotation and translation:
	\begin{equation}
	\begin{split}
	\mathbf{A} &= 
	\begin{bmatrix}
	\ p_{1}^1(\psi) & \cdots & p_{1}^1(\psi) \ \\
	\ \vdots & \ddots & \vdots \ \\
	\ p_{9}^9(\psi) & \cdots & p_{9}^9(\psi) \ \\
	\end{bmatrix}_{9 \times 9} , \\
	\mathbf{x} &= 
	\begin{bmatrix}
	t_x \ 
	t_y \
	t_z \
	d_{1x} \
	d_{1y} \
	d_{1z} \
	d_{2x} \
	d_{2y} \
	d_{2z} \
	\end{bmatrix}^{\top} .
	\end{split}
	\end{equation}

	\vspace{3mm}
	\noindent $\textbf{Angular 5-point algorithm.}$
	The angular rolling shutter camera model does not consider linear velocities.
	Thus, given the angular velocity measurement of the IMU, the rolling shutter epipolar constraint is expressed as:
	\begin{equation}
	\mathbf{m}_{2}^{\top} \mathbf{E}_{r} (\mathbf{R},\mathbf{t}; \mathbf{w}_1,\mathbf{w}_2) \mathbf{m}_{1} = 0 .
	\label{eq:rolling_shutter_epipolar_geometry}
	\end{equation}
	This angular rolling shutter essential matrix consists of 6 unknown variables like the essential matrix of the global shutter camera model (3+3+3+3-3-3=6); however, this is different from the global shutter epipolar geometry.

	Since translation is up to scale, at least 5 points are required for relative pose estimation.
	We reformulate this rolling shutter epipolar constraint as:
	\begin{equation}
	\begin{split}
	p_{1}^{i}(\mathbf{R}) t_x + p_{2}^{i}(\mathbf{R}) t_y +  p_{3}^{i}&(\mathbf{R}) t_z  =   0\\
	& \text{for} \ \ i = 1,2,\cdots,5.
	\end{split}
	\end{equation}
	We rearrange the above equation as a linear matrix equation that decouples relative rotation and translation as:
	\begin{equation}
	\begin{split}
	\mathbf{A} & = \begin{bmatrix}
	\ p_{1}^{1}(\mathbf{R}) & \cdots & p_{3}^{1}(\mathbf{R}) \ \\
	\vdots & \ddots & \vdots \ \\
	\ p_{1}^{5}(\mathbf{R}) & \cdots & p_{3}^{5}(\mathbf{R}) \ \\
	\end{bmatrix}_{5 \times 3} ,	
	\\
	\mathbf{x} & = 
	\begin{bmatrix}
	t_x \ 
	t_y \
	t_z \
	\end{bmatrix}^{\top} .
	\end{split}
	\end{equation}
	
	\vspace{3mm}
	\noindent $\textbf{Angular 3-point algorithm.}$
	Given gravity and angular velocity measurements, the angular rolling shutter epipolar constraint is expressed as:
	\begin{equation}
	\mathbf{m}_{2}^{\top} \mathbf{E}_{r} (\psi,\mathbf{t}; \mathbf{w}_1,\mathbf{w}_2,\mathbf{g}_1,\mathbf{g}_2) \mathbf{m}_{1} = 0 .
	\label{eq:rolling_shutter_epipolar_geometry}
	\end{equation}
	The rolling shutter essential matrix consists of 4 unknown variables (3+3+3+3-3-3-2=4).
	Since translation is up to scale, the relative pose can be estimated with minimum 3 points.
	This equation is reformulated as: 
	\begin{equation}
	\begin{split}
	p_{1}^{i}(\psi) t_x + p_{2}^{i}(\psi) t_y + & p_{3}^{i}(\psi) t_z =  0\\
	& \text{for} \ \ i = 1,2,3.
	\end{split}
	\end{equation}
	We stack the above equation as a linear matrix equation that decouples relative rotation and translation as:
	\begin{equation}
	\begin{split}
	\mathbf{A} & = \begin{bmatrix}
	\ p_{1}^{1}(\psi) & p_{2}^{1}(\psi) & p_{3}^{1}(\psi) \ \\
	\ p_{1}^{2}(\psi) & p_{2}^{2}(\psi) & p_{3}^{2}(\psi) \ \\
	\ p_{1}^{3}(\psi) & p_{2}^{3}(\psi) & p_{3}^{3}(\psi) \ \\
	\end{bmatrix}_{3 \times 3} , 
	\\
	\mathbf{x} & = 
	\begin{bmatrix}
	t_x \ 
	t_y \
	t_z \
	\end{bmatrix}^{\top} .
	\end{split}
	\end{equation}

	\vspace{3mm}
	\noindent $\textbf{Uniform 11-point algorithm.}$
	Since the uniform model takes both angular and linear velocities, given the angular velocity of the IMU, the rolling shutter epipolar constraint is written as:
	\begin{equation}
	{\mathbf{m}_{2}^{i}}^{\top} \mathbf{E}_{r} (\mathbf{R},\mathbf{t},\mathbf{d}_1,\mathbf{t}_2;  \mathbf{w}_1,\mathbf{w}_2) \mathbf{m}_{1}^{i} = 0 .
	\label{eq:rolling_shutter_epipolar_geometry}
	\end{equation}
	This rolling shutter essential matrix has 12 DOF (3+3+3+3 +3+3-3-3=12).
	As  previous algorithms, translation and linear velocity are up to scale, therefore, at least 11 points are needed to solve the problem.
	We reformulate this equation as:
	\begin{equation}
	\begin{split}
	& p_{1}^i(\mathbf{R}) t_x + p_{2}^i(\mathbf{R}) t_y + p_{3}^i(\mathbf{R}) t_z + p_{4}^i(\mathbf{R}) d_{1x} \\ 
	& + p_{5}^i(\mathbf{R}) d_{1y} + p_{6}^i(\mathbf{R}) d_{1z} + p_{7}^i(\mathbf{R}) d_{2x}  \\
	& + p_{8}^i(\mathbf{R}) d_{2y} + p_{9}^i(\mathbf{R}) d_{2z} =   0    \ \ \ \ \ \ \ \ \ \ \ \text{for} \ \ i = 1,2,\cdots,11.
	\end{split}
	\end{equation}
	We stack the above equation as a linear matrix equation that decouples relative rotation and translation as:
	\begin{equation}
	\begin{split}
	\mathbf{A} &= 
	\begin{bmatrix}
	\ p_{1}^1(\mathbf{R}) & \cdots & p_{1}^1(\mathbf{R}) \ \\
	\ \vdots & \ddots & \vdots \ \\
	\ p_{9}^{11}(\mathbf{R}) & \cdots & p_{9}^{11}(\mathbf{R}) \ \\
	\end{bmatrix}_{11 \times 9} , 
	\\
	\mathbf{x} &= 
	\begin{bmatrix}
	t_x \ 
	t_y \
	t_z \
	d_{1x} \
	d_{1y} \
	d_{1z} \
	d_{2x} \
	d_{2y} \
	d_{2z} \
	\end{bmatrix}^{\top} .
	\end{split}
	\end{equation}

	\vspace{3mm}
	\noindent $\textbf{Uniform 9-point algorithm.}$
	Given the gravity and angular velocity of the IMU, the rolling shutter epipolar geometry is given as: 
	\begin{equation}
	{\mathbf{m}_{2}^{i}}^{\top} \mathbf{E}_{r} (\psi,\mathbf{t},\mathbf{d}_1,\mathbf{d}_2  ; \mathbf{w}_1, \mathbf{w}_2, \mathbf{g}_1, \mathbf{g}_2)  \mathbf{m}_{1}^{i} = 0 .
	\label{eq:rolling_shutter_epipolar_geometry}
	\end{equation}
	This rolling shutter essential matrix  has 10 unknown variables, and at least 9 points are required to estimate relative pose due to the scale ambiguity of translation and linear velocity.
	This equation is reformulated as:
	\begin{equation}
	\begin{split}
	& p_{1}^i(\psi) t_x + p_{2}^i(\psi) t_y + p_{3}^i(\psi) t_z + p_{4}^i(\psi) d_{1x} \\ 
	& + p_{5}^i(\psi) d_{1y} + p_{6}^i(\psi) d_{1z} + p_{7}^i(\psi) d_{2x} \\
	& + p_{8}^i(\psi) d_{2y} + p_{9}^i(\psi) d_{2z} =   0    \  \ \ \ \ \ \ \ \ \ \  \text{for} \ \ i = 1,2,\cdots,9.
	\end{split}
	\end{equation}
	We stack these equations as a matrix as:
	\begin{equation}
	\begin{split}
	\mathbf{A} &= 
	\begin{bmatrix}
	\ p_{1}^1(\psi) & \cdots & p_{1}^1(\psi) \ \\
	\ \vdots & \ddots & \vdots \ \\
	\ p_{9}^9(\psi) & \cdots & p_{9}^9(\psi) \ \\
	\end{bmatrix}_{9 \times 9} ,
	\\
	\mathbf{x} &= 
	\begin{bmatrix}
	t_x \ 
	t_y \
	t_z \
	d_{1x} \
	d_{1y} \
	d_{1z} \
	d_{2x} \
	d_{2y} \
	d_{2z} \
	\end{bmatrix}^{\top} .
	\end{split}
	\end{equation}

	\subsection{Solver} \label{subsec:solver}

	From the homogeneous equation $\mathbf{A}\mathbf{x}=\mathbf{0}$ constructed in each algorithm, we first estimate the relative rotation $\mathbf{R}$, and then estimate the relative translation $\mathbf{t}$.
	For the estimation of rotation, we use the rank deficiency of $\mathbf{A}$.
	Since $\mathbf{x}$ is determined up to scale, the rank of $\mathbf{A}$ is always equal to $dim(\mathbf{x})-1$
	Therefore, the determinant of $\mathbf{A}$ must be 0.
	We estimate the unknown variable $ \psi $ or $ \mathbf{R} $ through the Levenberg-Marquart algorithm to minimize the determinant(s):
	\begin{equation}
	\begin{split}
	\min_{\psi}  & {
		\left| \mathbf{A}(\psi) \right|
	},
	\\
	\min_{\mathbf{R}}  & {
		\left| \mathbf{A}_{1}(\mathbf{R}) \right|  +  \left| \mathbf{A}_{2}(\mathbf{R}) \right| + \left| \mathbf{A}_{3}(\mathbf{R}) \right|
	},
	\end{split}
	\end{equation}
	where $|\cdot|$ denotes the determinant of the matrix.
	$\mathbf{A}_1$, $\mathbf{A}_2$, and $\mathbf{A}_3$ are $3 \times 3$ submatrices of A in the angular 5-point, and $9 \times 9$ submatrices of $\mathbf{A}$ in the uniform 11-point algorithm.
	We parameterize $\mathbf{R}$ to unit-quaternion $\mathbf{q} \in \mathbb{R}^{4}$ because the quaternion has lower dimension compared to $\mathbf{R}$:
	\begin{equation}
	\begin{split}
	\min_{\mathbf{q}}  & {
		\left| \mathbf{A}_{1}(\mathbf{q}) \right|  +  \left| \mathbf{A}_{2}(\mathbf{q}) \right| + \left| \mathbf{A}_{3}(\mathbf{q}) \right| .
	}
	\end{split}
	\end{equation}
	
	The optimization on the quaternion is implemented through  the following local parameterization as 
	\begin{equation}
	\odot(\mathbf{q}, \Delta \mathbf{q}) = \left[ \cos(|\Delta \mathbf{q}|), \frac{ \sin(|\Delta \mathbf{q}|)}{|\Delta \mathbf{q}|} \Delta \mathbf{q} \right] \ast  \mathbf{q}  ,
	\end{equation}
	where $ \odot $ is an addition operation, and $\Delta \mathbf{q}$ is a three-dimensional vector.
	For initialization of $\psi$, we use the estimate of the 5-point algorithm of the global shutter camera model.

	After estimating relative rotation, we estimate $\mathbf{x}$ using singular value decomposition (SVD).
	The estimated $ \mathbf{x} $ has two solutions  as $ [\mathbf{t}, \ \mathbf{d}_1, \ \mathbf{d}_2]$ and $[-\mathbf{t}, \ -\mathbf{d}_1, \ -\mathbf{d}_2] $.

	\subsection{Refinement} \label{subsec:refinement}
	
	We refine the estimate from the linear algorithm with the inlier points obtained through RANSAC.
	To this end, we define the following energy minimization problem as
	\begin{equation}
	\begin{split}
	\min_{\mathbf{q},\mathbf{t},\mathbf{d}_1,\mathbf{d}_2 }&{E(\mathbf{q},\mathbf{t},\mathbf{d}_1,\mathbf{d}_2 )^2 }+ \lambda_{t}(||\mathbf{t}||_{2}-1)^2\\
	& + \lambda_{d_1}||\mathbf{d}_1||_{2}^{2} + \lambda_{d_2}||\mathbf{d}_2||_{2}^{2} ,
	\end{split}
	\end{equation}
	where $ \mathbf{q} \in \mathbb{R}^{4}$ is the unit-quaternion representing the relative rotation, and $\lambda_{t}$, $\lambda_{d_1}$ and $\lambda_{d_2}$ are coefficients for regularization.
	%
	

	The energy function $E$ is defined by the well-known Sampson error:
	{
		\small
		\begin{equation}
		\begin{split}
		& E(\mathbf{q},\mathbf{t},\mathbf{d}_1,\mathbf{d}_2)  =  \\ 
		& \frac{\mathbf{m}_{2}^{\top}\mathbf{E}_{rs}\mathbf{m}_{1}} {\sqrt{(\mathbf{m}_{2}^{\top}\mathbf{E}_{rs})_{0}^2+(\mathbf{m}_{2}^{\top}\mathbf{E}_{rs})_{1}^2+(\mathbf{E}_{rs}\mathbf{m}_{1})_{0}^2+(\mathbf{E}_{rs}\mathbf{m}_{1})_{1}^2}} .
		\end{split}
		\end{equation}
	}

	Since the relative translation $ \mathbf{t} $ is the unit-vector, we add the constraint on $\mathbf{t}$ to the energy function, and $||\mathbf{d}_1||_2^2$ and $||\mathbf{d}_2||_2^2$ are regularization terms.

	The rolling shutter essential matrix is decoupled with $\mathbf{q}$ and $[\mathbf{t}, \mathbf{d}_1, \mathbf{d}_2]^{\top}$ in the proposed algorithms.
	Therefore, using these properties, we apply an alternative directing scheme  that estimates unknowns by alternating two parts until convergence.
	
	\section{Experiments} 
	
	We evaluated the performance of the proposed methods on both synthetic and real data experiments.
	We compare the performance of the global shutter 5-point algorithm \cite{Kneip:ICRA:2014,Nister:PAMI:2004} with the performance of the proposed uniform 9- and 11-point algorithms.\footnote{In this paper, we evaluate only the uniform 9- and 11-point algorithms among proposed five algorithms, since the uniform 9- and 11-point algorithms are more general algorithms that can be applied with less assumptions and limitations.}
	The evaluation metric for rotation is defined as $\arccos((\text{trace}(\mathbf{R}_{\text{gt}}^{-1}\mathbf{R}_{\text{est}})-1)/2) $ and for translation it is defined as $\arccos(\mathbf{t}_{\text{gt}}^{\top} \mathbf{t}_{\text{est}}) $,
	where $\mathbf{t}$ is a unit vector.

	\subsection{Synthetic Data}

	We randomly generate 3D points, positions, and orientations of two cameras for each evaluation.
	The number of the 3D points is about $300$ and the average distance of them from the cameras is about $20m$ .
	We set the positions and orientations of two cameras so that two cameras have forward and sideways motion, and the distance between two cameras is about $2m$ on average.
	Each camera 
	has an orientation within a range of 20$^{\circ}$ for each axis.
	The linear velocities $\mathbf{d}_1,\mathbf{d}_2$ and angular velocities $\mathbf{w}_1, \mathbf{w}_2$ are manually set in accordance with the experimental purpose.
	The rolling shutter readout time $\lambda_{r}$ is set to $60 us$.
	The image resolution is set to $1920 \times 1080$, focal length is set to $640$ pixel, and radial distortion is not considered.
	The generated points are projected onto the image plane of each camera with the given intrinsic camera parameters and rolling shutter parameters (angular, linear velocities, and readout time).
	We remove the points that are out of the field of view or have no corresponding points.
	Inertial measurements are generated by transforming the gravity and the angular velocity to the IMU coordinate system through the extrinsic parameters between the given IMU and the camera.
	The gravity in the world coordinate $\mathbf{g}_{w}$ is set to $[0 \ 0 \ 9.81]^{\top}$.
	We repeat all the experiments 100 times to obtain statistically meaningful results.
	
	\begin{figure}[t]
		\begin{subfigure}[b]{0.99\linewidth}
			\includegraphics[width=\linewidth]{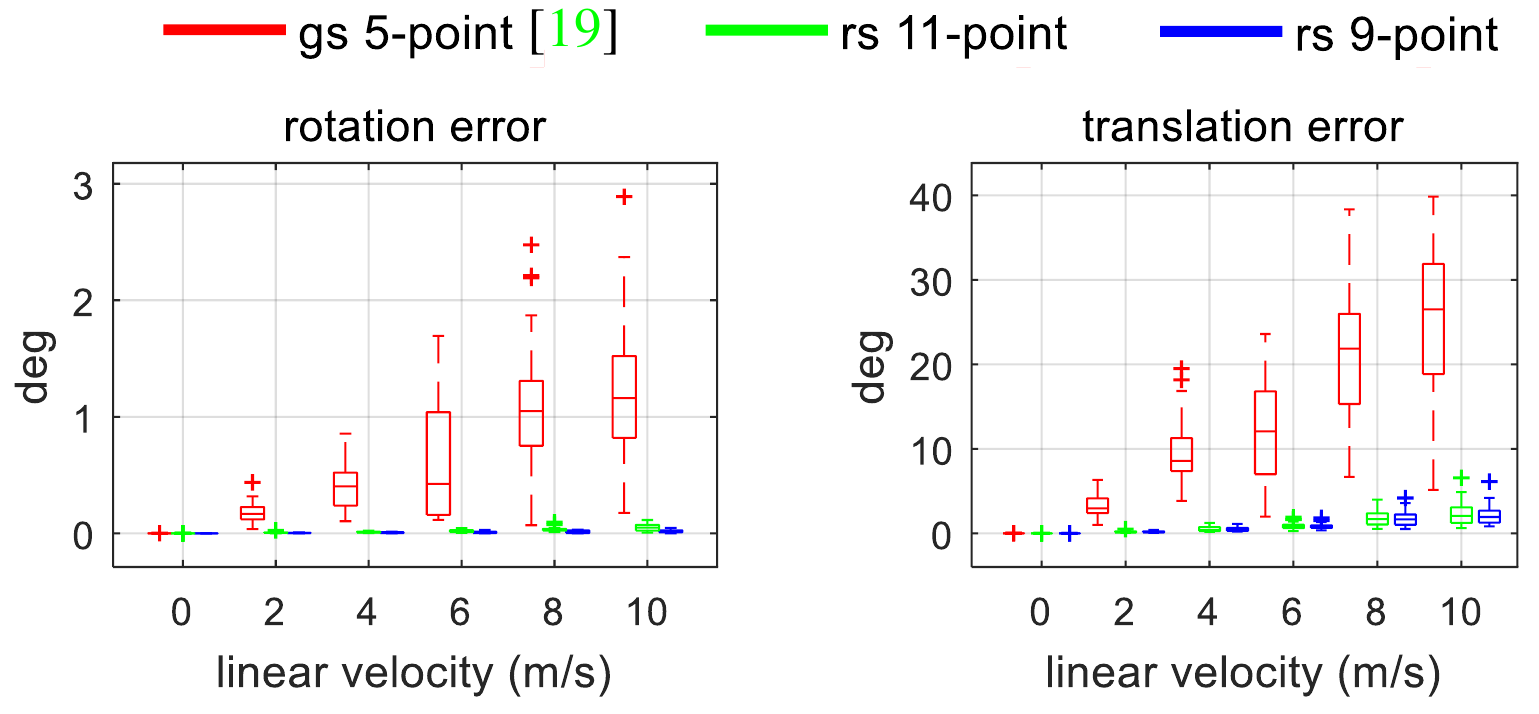}
			\caption{Forward motion case}
		\end{subfigure}
		\begin{subfigure}[b]{0.99\linewidth}
			\includegraphics[width=\linewidth]{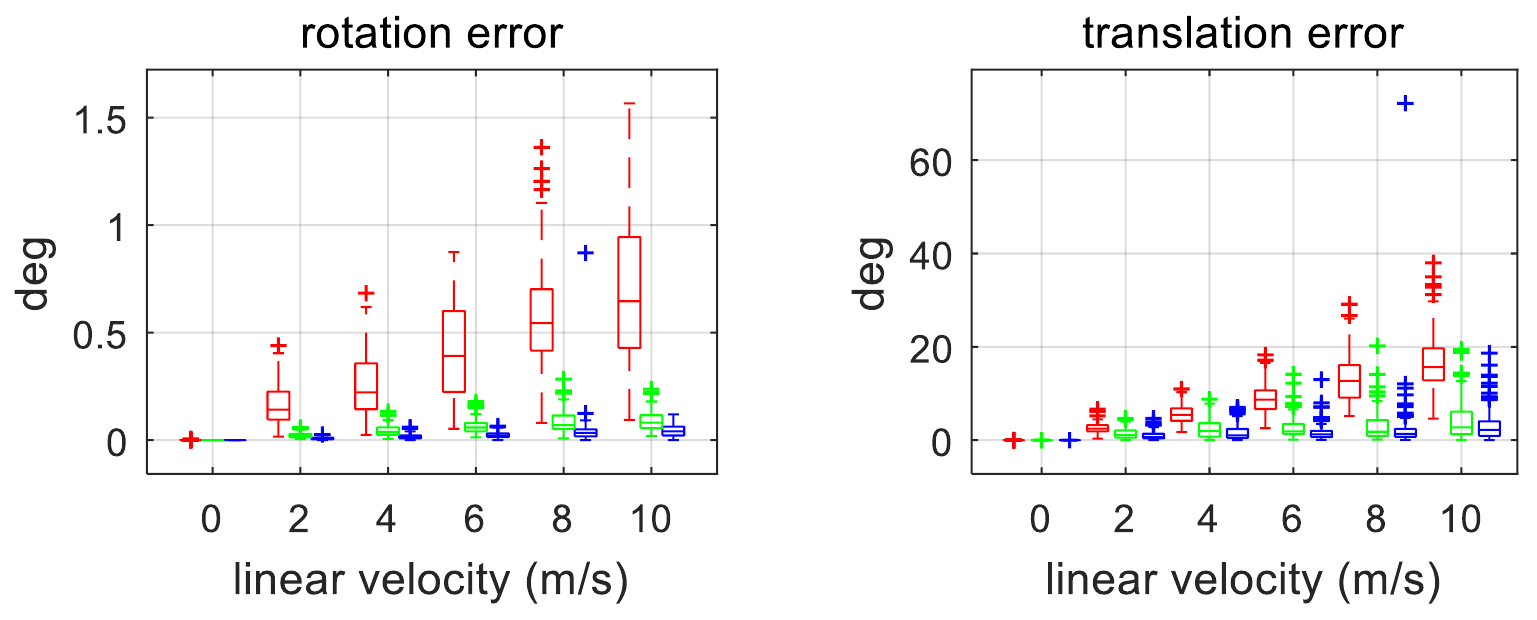}
			\caption{Sideways motion case}
		\end{subfigure}%
		\caption{Performance comparison with increasing linear velocity}
		\label{fig:exp_synthetic_rs_d}		
	\end{figure}
	
	\begin{figure}[t]
		\begin{subfigure}[b]{0.99\linewidth}
			\includegraphics[width=\linewidth]{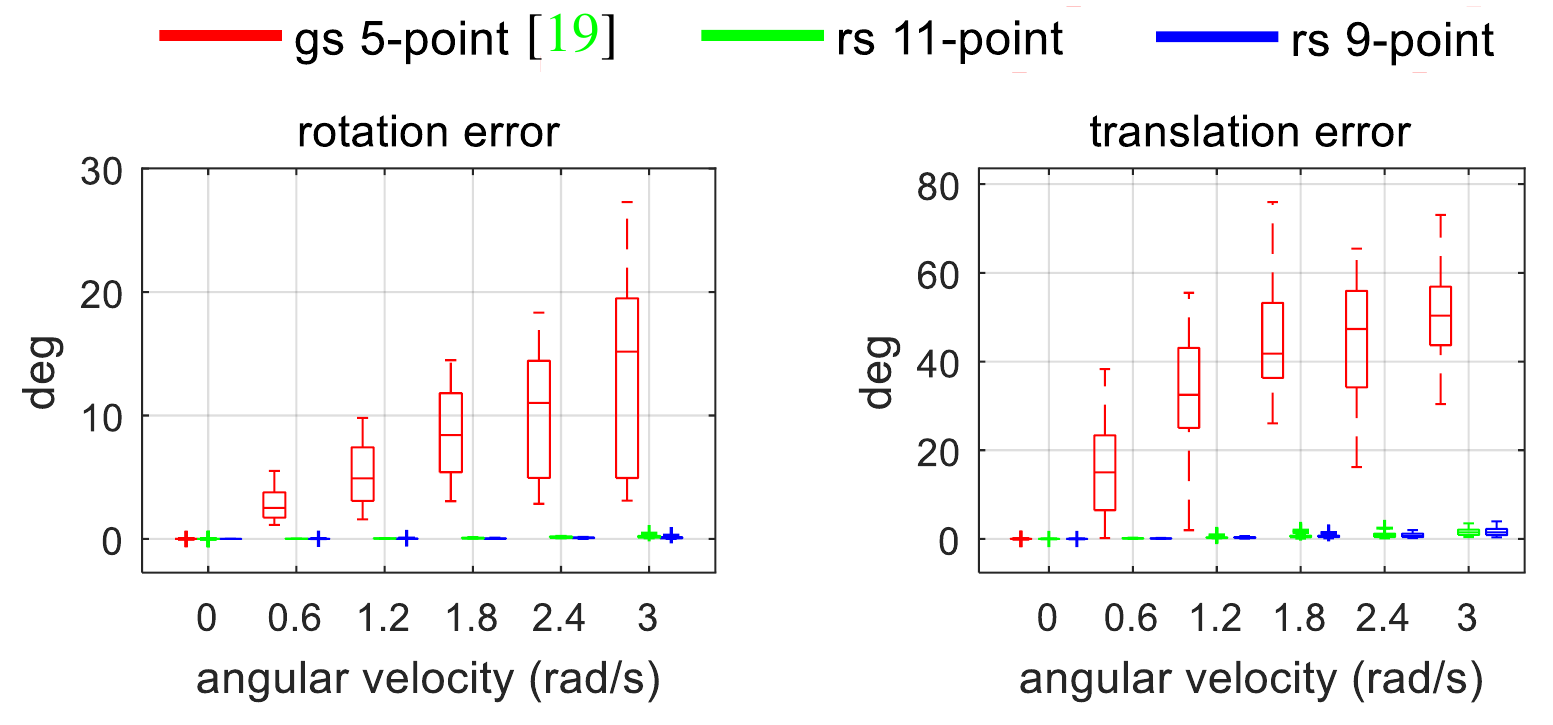}
			
			\caption{Forward motion case}
		\end{subfigure}
		\begin{subfigure}[b]{0.99\linewidth}
			\includegraphics[width=\linewidth]{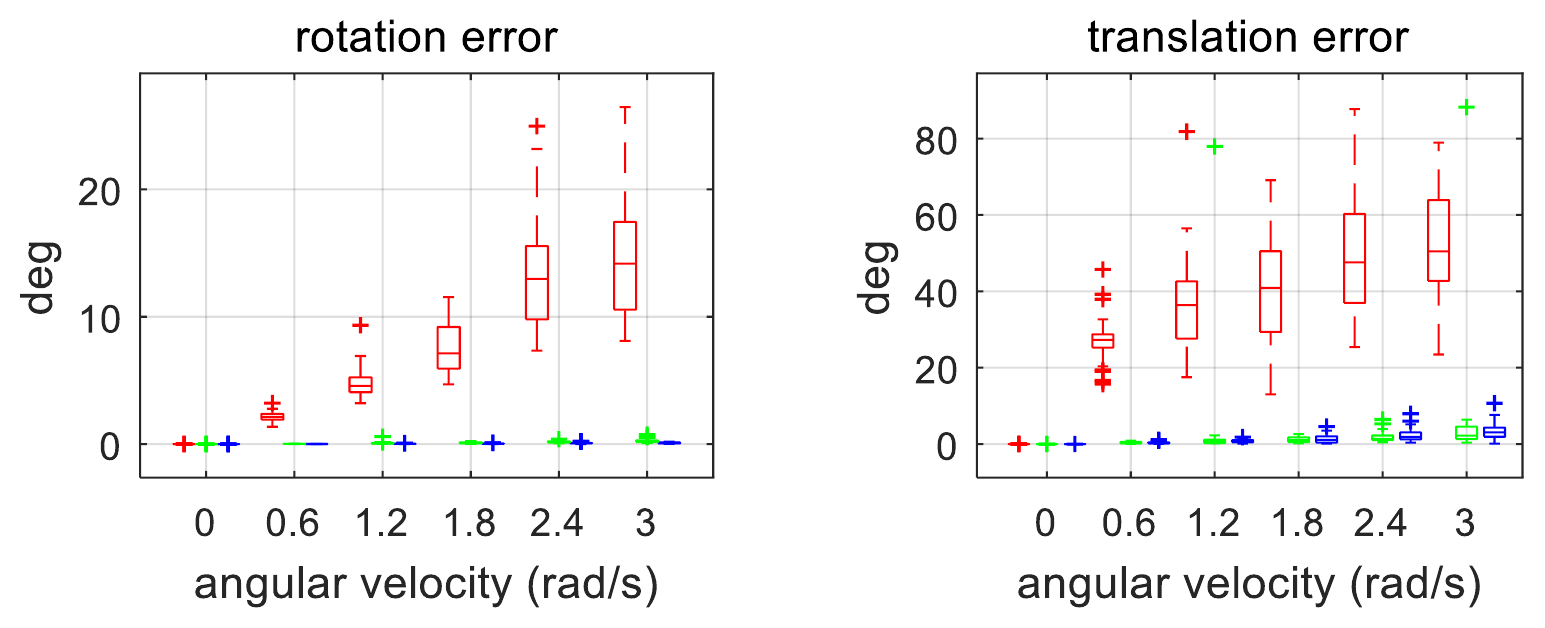}
			\caption{Sideways motion case}
		\end{subfigure}%
		\caption{Performance comparison with increasing angular velocity}
		\label{fig:exp_synthetic_rs_w}
	\end{figure}

	\begin{figure}[t]
		\begin{subfigure}[b]{0.99\linewidth}
			\includegraphics[width=\linewidth]{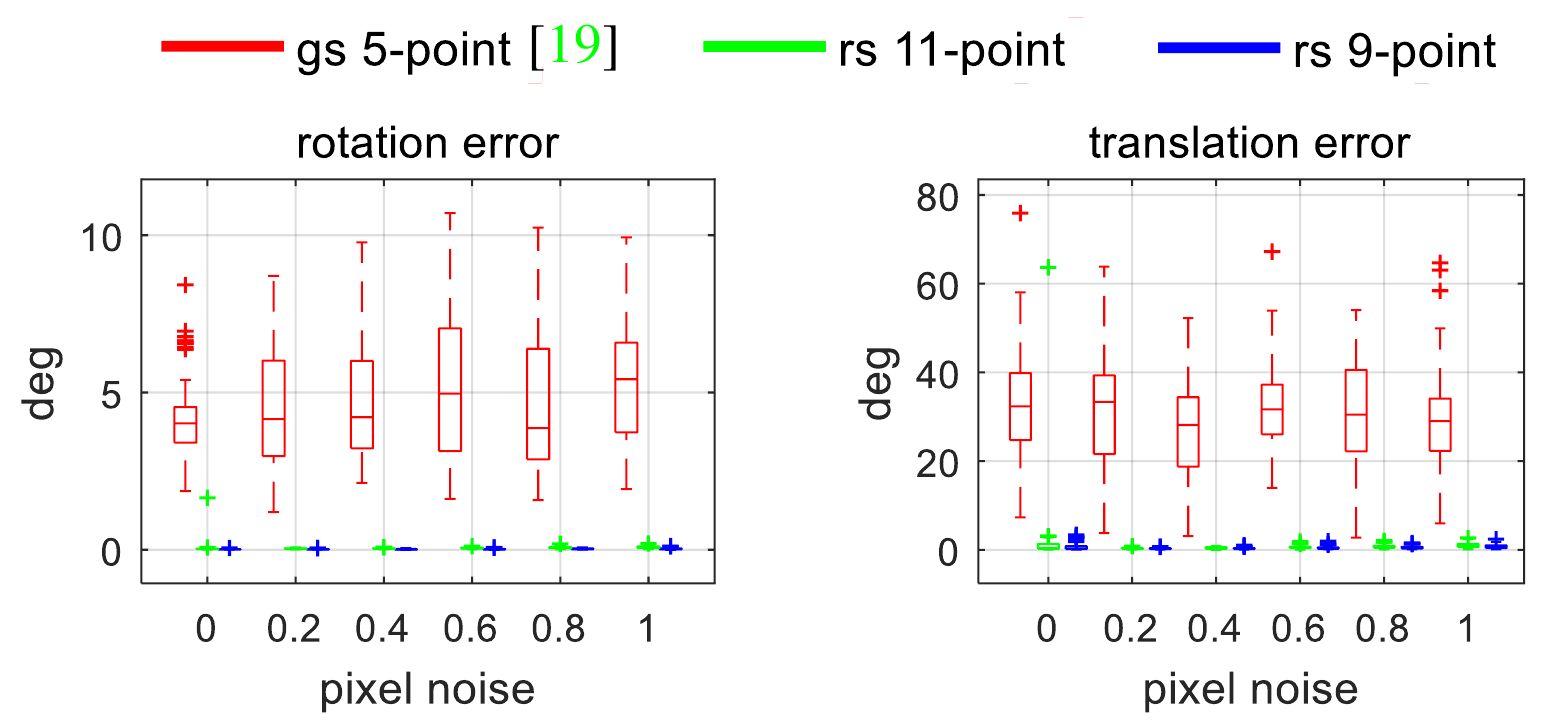}
			\caption{Forward motion case}
		\end{subfigure}
		\begin{subfigure}[b]{0.99\linewidth}
			\includegraphics[width=\linewidth]{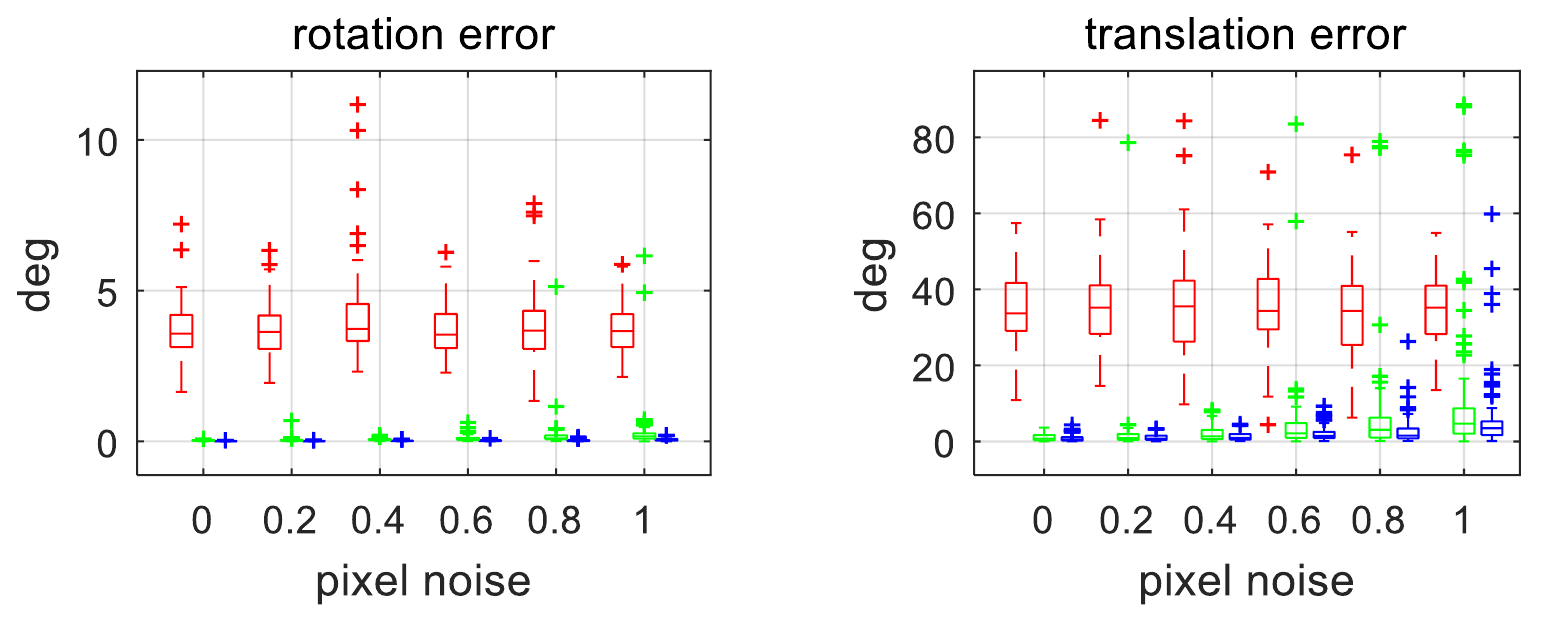}
			\caption{Sideways motion case}
		\end{subfigure}%
		\caption{Performance evaluation with increasing feature point noise.}
		\label{fig:exp_synthetic_pixel_noise}
	\end{figure}

	First, we evaluate each algorithm in the presence of instantaneous camera motion and the absence of noise.
	To analyze the effects of linear velocity and angular velocity, we perform experiments with increasing linear velocity and zero angular velocity.
	Then, we perform experiments with increasing angular velocity and zero linear velocity.
	The magnitude of the linear velocity is increased from $0$ to $10 \ m/s$ for each axis.
	Since the relative pose estimation is for an unordered image pair, we set the linear velocity in two cameras to the opposite direction.
	Figure~\ref{fig:exp_synthetic_rs_d} shows that the rotation and translation errors of the 5-point algorithm \cite{Nister:PAMI:2004} increase significantly compared to the proposed 9- and 11-point algorithms as the linear velocity increases.
	The proposed algorithms maintain rotation errors less than 0.1$^{\circ}$ in both forward and sideways motion  cases even if the linear velocity changes.
	Interestingly, the translation estimates of the proposed algorithms are more accurate in the forward motion case than in the sideways motion case, as opposed to the 5-point algorithm.
	The magnitude of the angular velocity increases from $0$ to $3 \ rad/s$ for each axis, and is set reversely for both cameras as for the linear velocity.
	Figure~\ref{fig:exp_synthetic_rs_w}  shows that the proposed algorithms for both motions are much more accurate than the 5-point algorithm.
	The increase of the angular velocity further degrades relative pose estimation performance rather than the increase of linear velocity, but the proposed method is not significantly affected.

	\begin{figure}[t]
		\begin{subfigure}[b]{0.99\linewidth}
			\includegraphics[width=\linewidth]{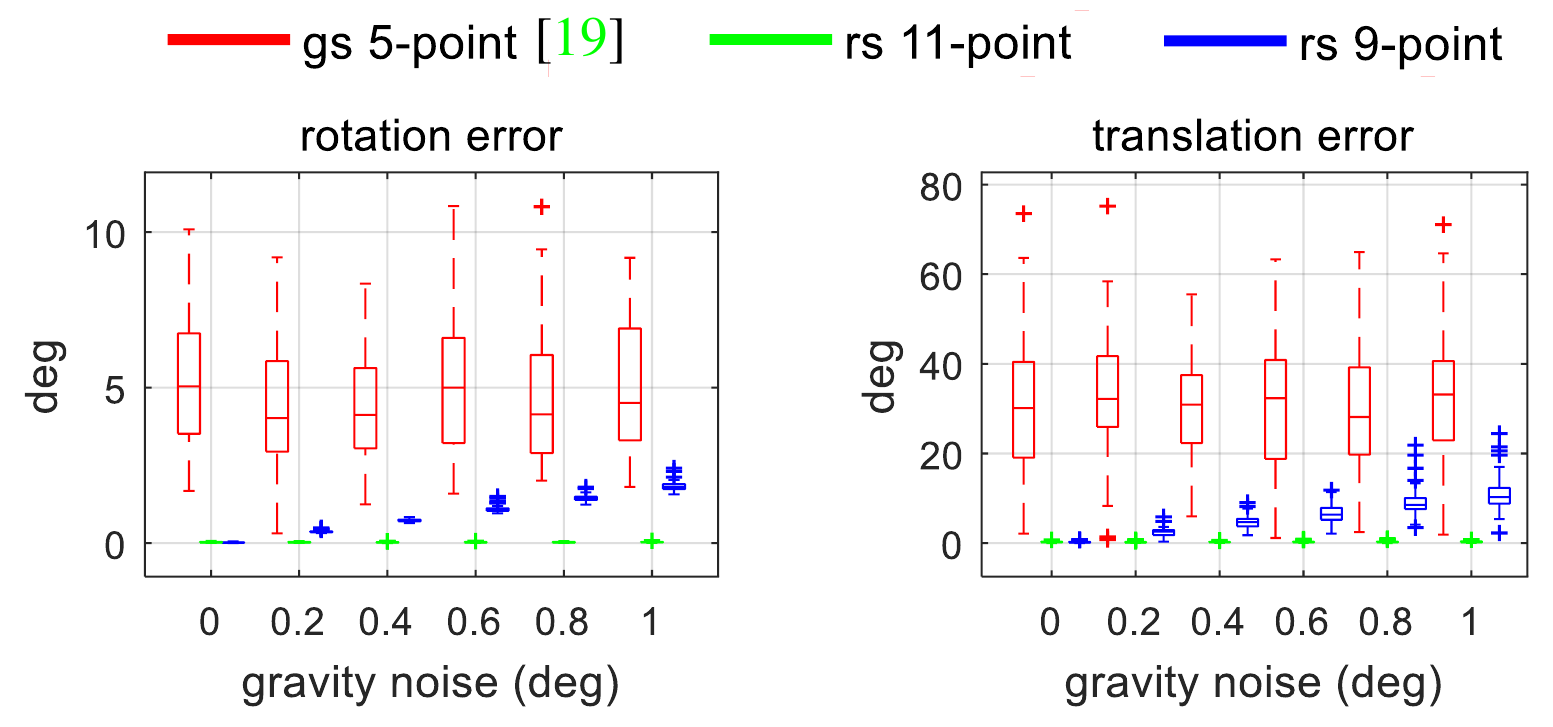}
			\caption{Forward motion case}
		\end{subfigure}
		\begin{subfigure}[b]{0.99\linewidth}
			\includegraphics[width=\linewidth]{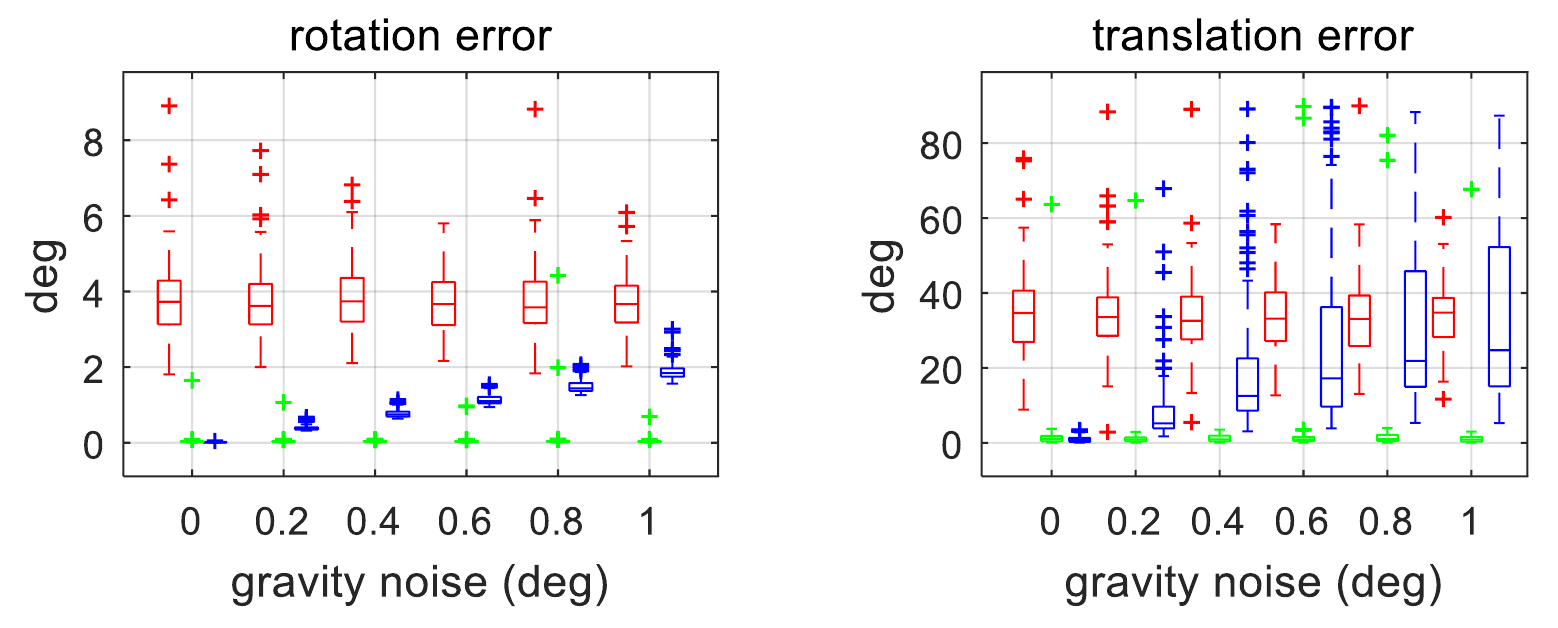}
			\caption{Sideways motion case}
		\end{subfigure}%
		\caption{Performance evaluation with increasing gravity measurement noise.}
		\label{fig:exp_synthetic_g_noise}
	\end{figure}

	Next, we perform experiments with increasing noise of corresponding points in the image.
	The linear velocity is set to 1 $m/s$ and the angular velocity to 1 $rad/s$.
	These values were determined by observing the dataset used in our real data experiments.
	The standard deviation of the noise increases from 0 to 1 pixel.
	Figure~\ref{fig:exp_synthetic_pixel_noise} shows that the point noise does not significantly affect the performance of the rolling shutter relative pose estimation in forward motion cases.
	In sideways motion cases, the outliers in translaion estimates of the proposed algorithms increase.

\begin{figure}[t]
		\begin{subfigure}[b]{0.99\linewidth}
			\includegraphics[width=\linewidth]{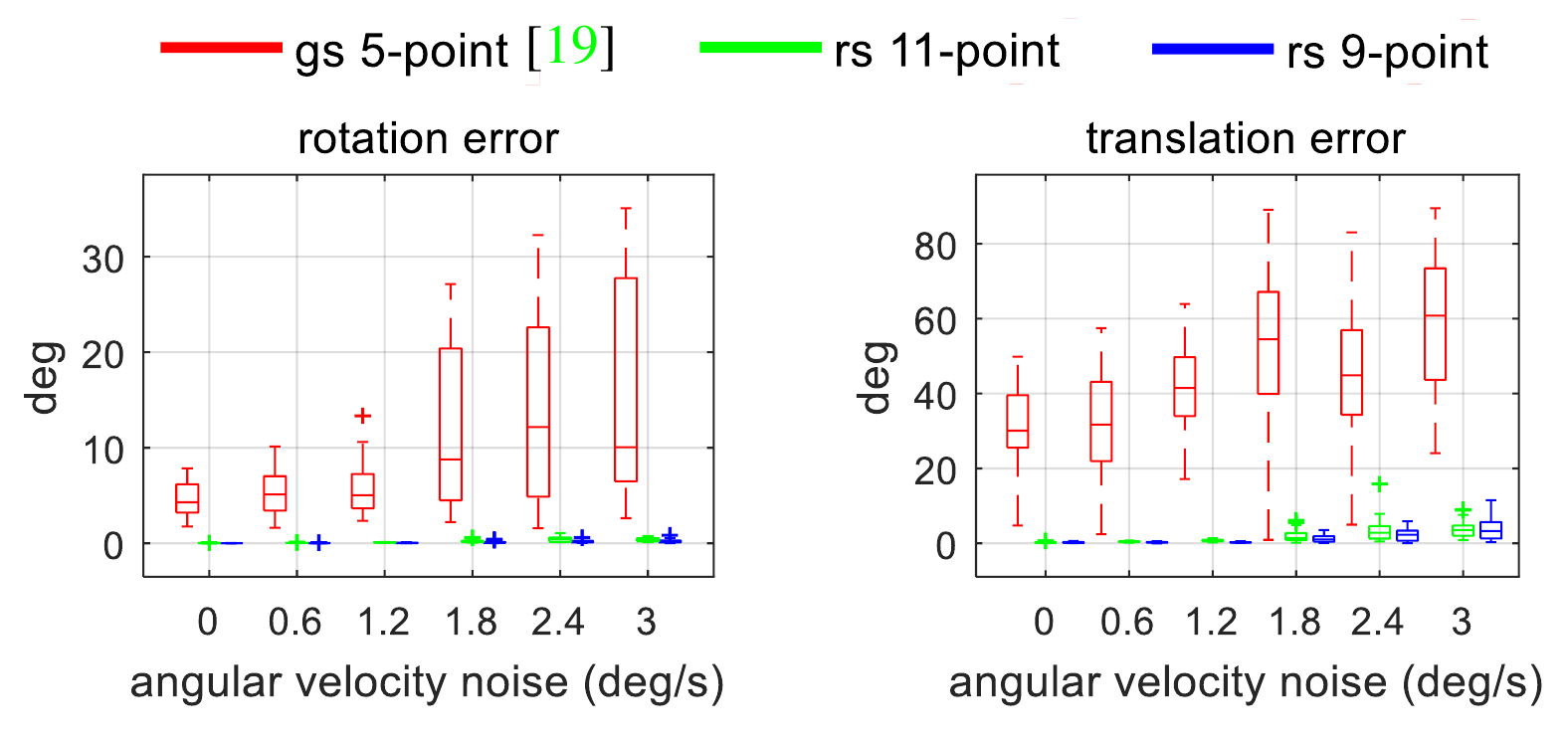}
			\caption{Forward motion case}
		\end{subfigure}
		\begin{subfigure}[b]{0.99\linewidth}
			\includegraphics[width=\linewidth]{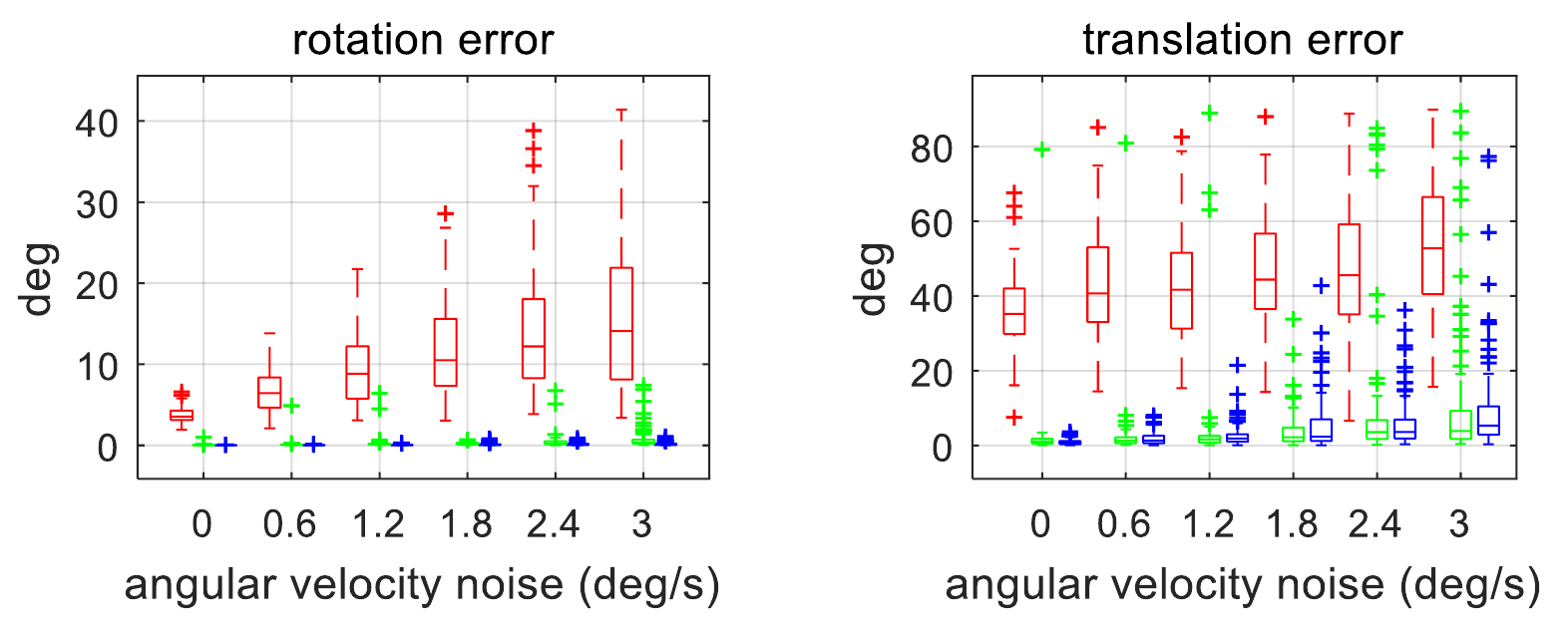}
			\caption{Sideways motion case}
		\end{subfigure}%
		\caption{Performance evaluation with increasing angular velocity measurement noise.}
		\label{fig:exp_synthetic_w_noise}
	\end{figure}

	The performance of the proposed algorithms is dependent on the accuracy of the inertial measurements.
	Thus, we analyze the performance against the noise increase of the inertial measurement.
	The IMU's gravity measurement error is less than 0.5$^{\circ}$ for low-cost MEMS sensors and less than 0.01$^{\circ}$ for high-priced sensors.
	We compare the performance while increasing the noise of the gravity up to 1$^{\circ}$ for each axis considering the camera motion.
	The gravity noises in the two camera frames are applied in the opposite direction.
	As before, the linear and angular velocities are set to 1 $m/s$ and to 1 $rad/s$ respectively.
	Figure~\ref{fig:exp_synthetic_g_noise} shows that the performance of the proposed 9-point algorithm decreases as gravity noise increases.
	This performance degradation becomes more severe in the sideways motion case than in the forward motion case.
	On the other hand, the proposed 11-point algorithm can be used even if the gravity noise is large.
	We also perform experiments with increasing noise of the angular velocity measurement.
	The angular velocity noise is less than 0.1$^{\circ}$ deg even for the low-cost sensor.
	We compare the performance by increasing the noise up to 3$^{\circ}$ assuming severe noise.
	Figure~\ref{fig:exp_synthetic_w_noise} shows that the angular velocity noise does not significantly affect the performance of the relative pose estimation in the forward motion case.
	However, in the sideways motion case, the increasing angular velocity noise causes some outliers in the translation estimates of the proposed algorithms.

	\subsection{Real Data}

	\begin{figure}[t]
		\centering	
		\begin{subfigure}[b]{0.49\linewidth}
			\includegraphics[width=\linewidth]{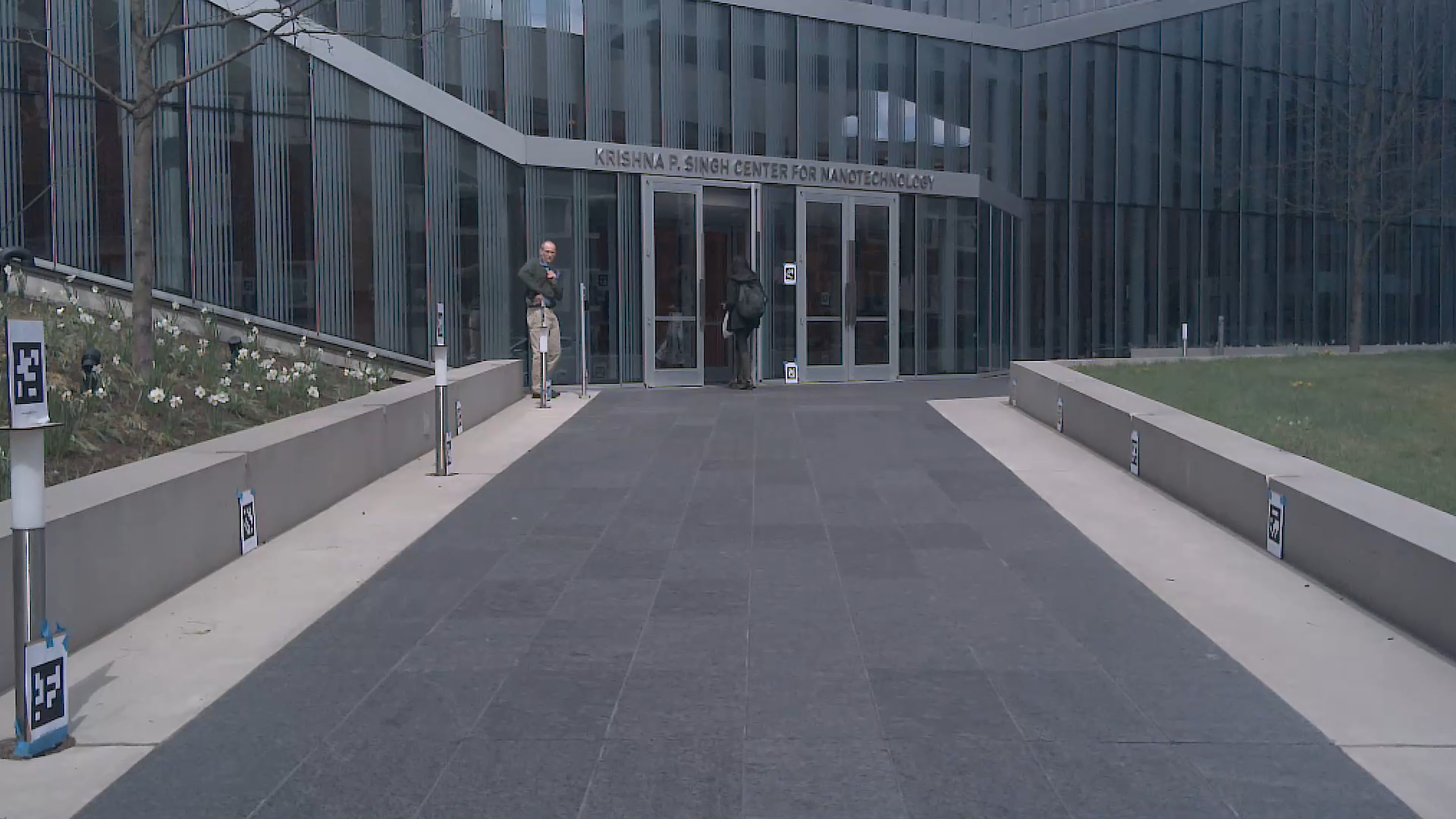}
			\caption{frame 393 (as-1)}
		\end{subfigure}
		\begin{subfigure}[b]{0.49\linewidth}
			\includegraphics[width=\linewidth]{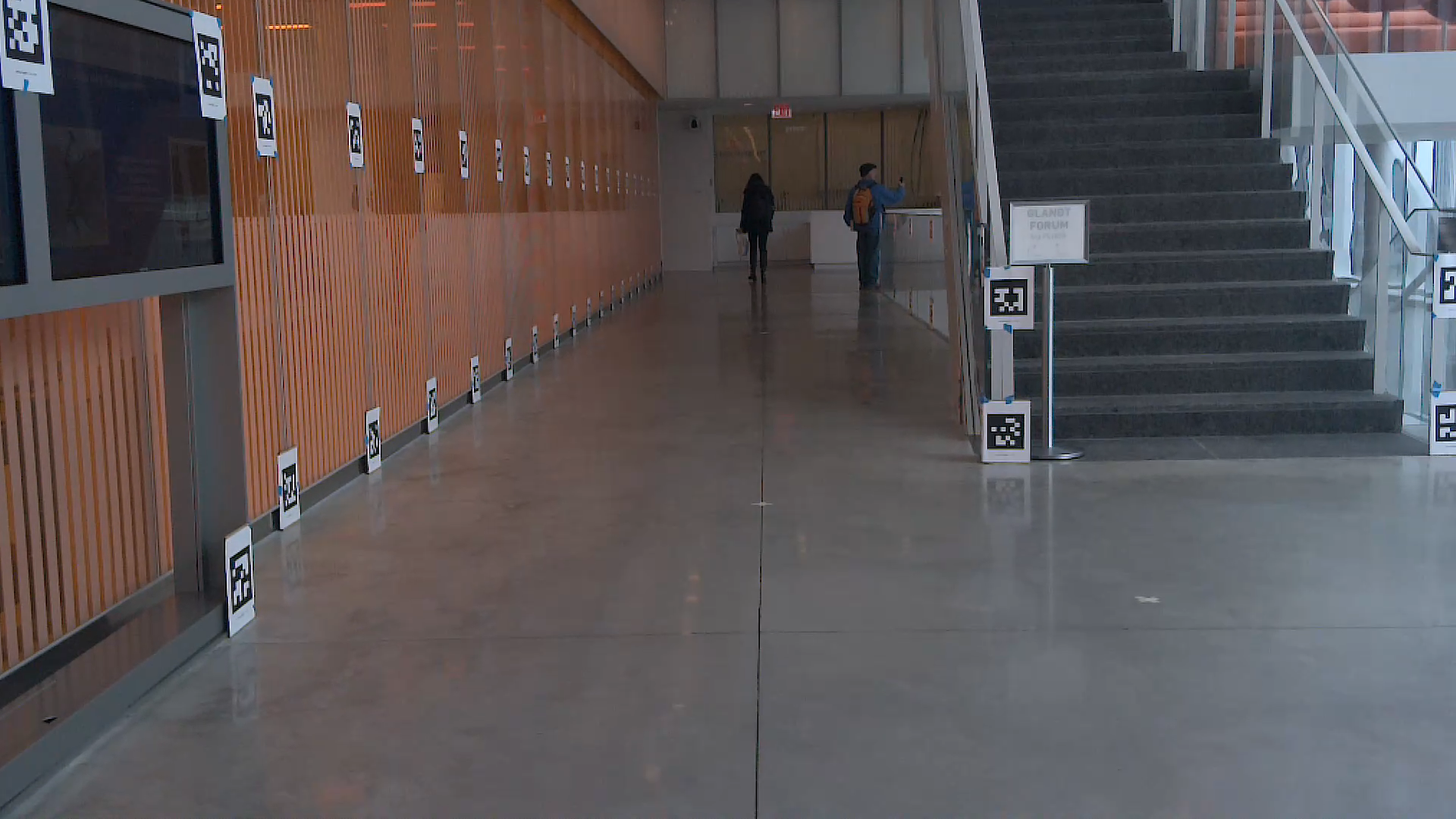}
			\caption{frame 1373 (as-2)}
		\end{subfigure}\\ \vspace{2mm}
		\begin{subfigure}[b]{0.49\linewidth}
			\includegraphics[width=\linewidth]{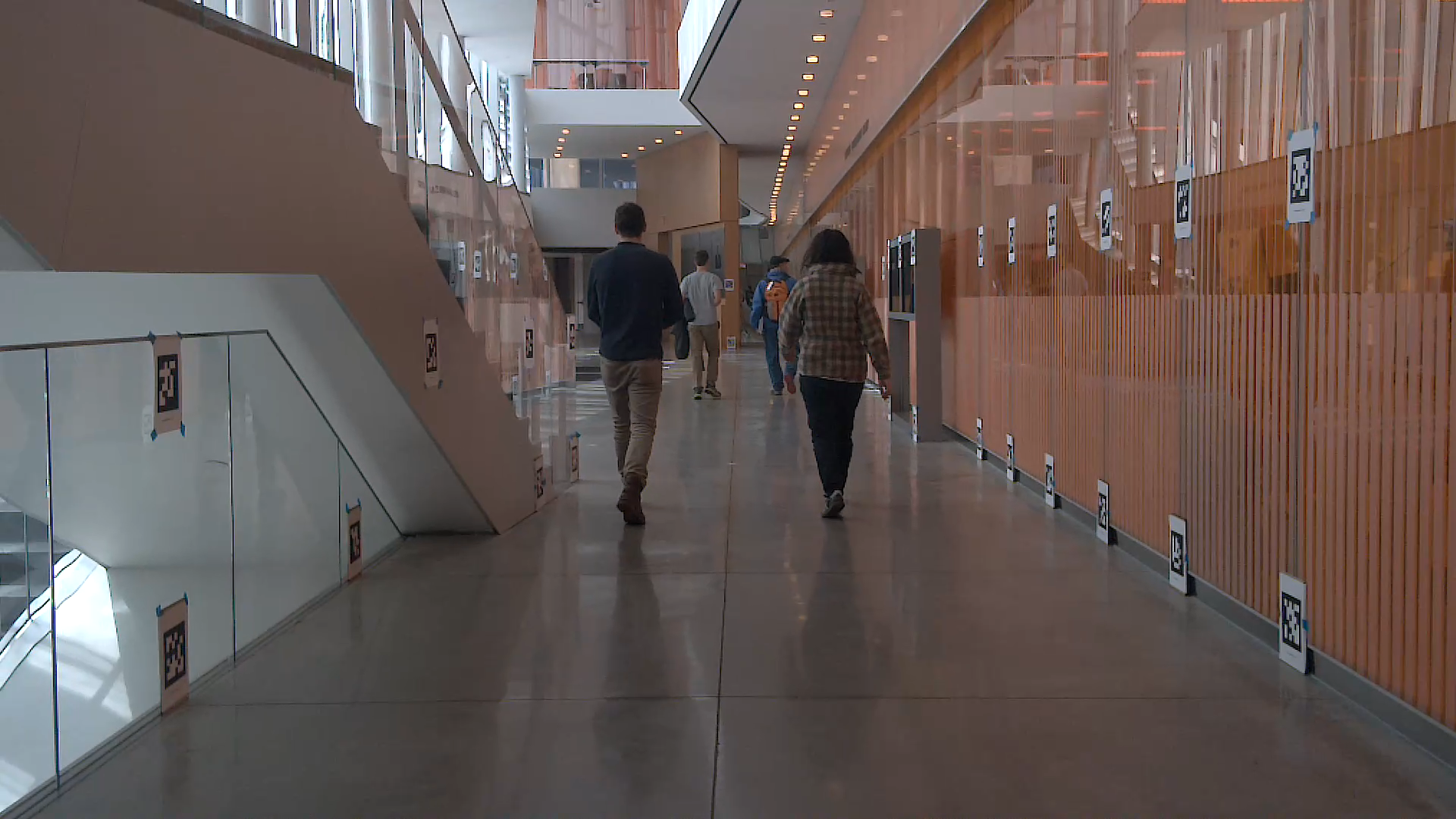}
			\caption{frame 2706 (as-3)}
		\end{subfigure}
		\begin{subfigure}[b]{0.49\linewidth}
			\includegraphics[width=\linewidth]{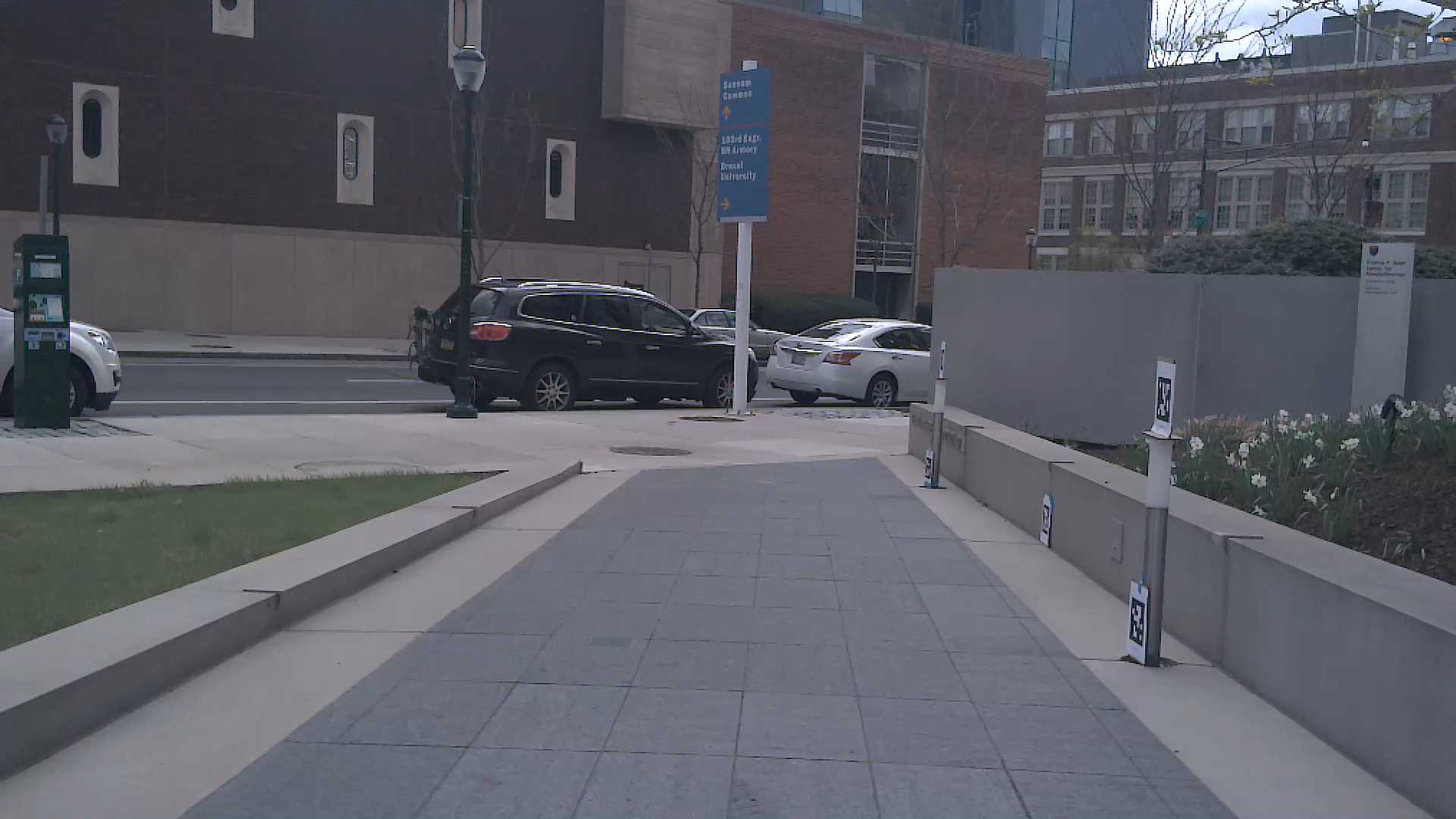}
			\caption{frame 4461 (as-4)}
		\end{subfigure}
		\caption{Sample images of the `as' sequence used in the real data experiments.}		
		\label{fig:real_data_sample_images} 
	\end{figure}
	
	\begin{table}[t]
		\setlength{\tabcolsep}{3pt}
		\footnotesize
		\caption{Detailed information on the real dataset used for experiments.} 
		\label{table:real_data_details}
		\begin{tabular}{|c|c|c|c|c|c|}
			\hline
			\multicolumn{2}{|c|}{sequence}             & frame range & \begin{tabular}[c]{@{}c@{}} $\#$ of image \\ pairs \end{tabular} & \begin{tabular}[c]{@{}c@{}} linear velocity\\ (m/s)\end{tabular} & \begin{tabular}[c]{@{}c@{}} angular velocity\\ (deg/s)\end{tabular}   \\ \hline \hline
			\multirow{4}{*}{as} & 1 & 206 - 567  & 67  & 1.31                & 11.76               \\ \cline{2-6}
			& 2 & 1319 -  1679 & 66 & 1.16                & 13.33                \\ \cline{2-6}
			& 3 & 2581 -  2971 & 72 & 1.08                & 12.30                \\ \cline{2-6}
			& 4 & 4358 -  4599 & 43 & 1.41                & 15.00                \\ \hline \hline
			\multirow{4}{*}{af} & 1  & 101 - 551 & 86  &  2.08                & 15.17                \\ \cline{2-6}
			& 2    & 913 - 1123 & 38  &  1.88                & 19.66              \\ \cline{2-6}
			& 3   & 1604 - 1815 & 39  &  2.00                & 18.77              \\ \cline{2-6}
			& 4   & 2539 - 2810 & 51  &  2.36                & 16.96               \\ \hline
		\end{tabular}
	\end{table}

	\begin{figure}[t]		
		\begin{subfigure}[b]{0.99\linewidth}					
			\includegraphics[width=\linewidth]{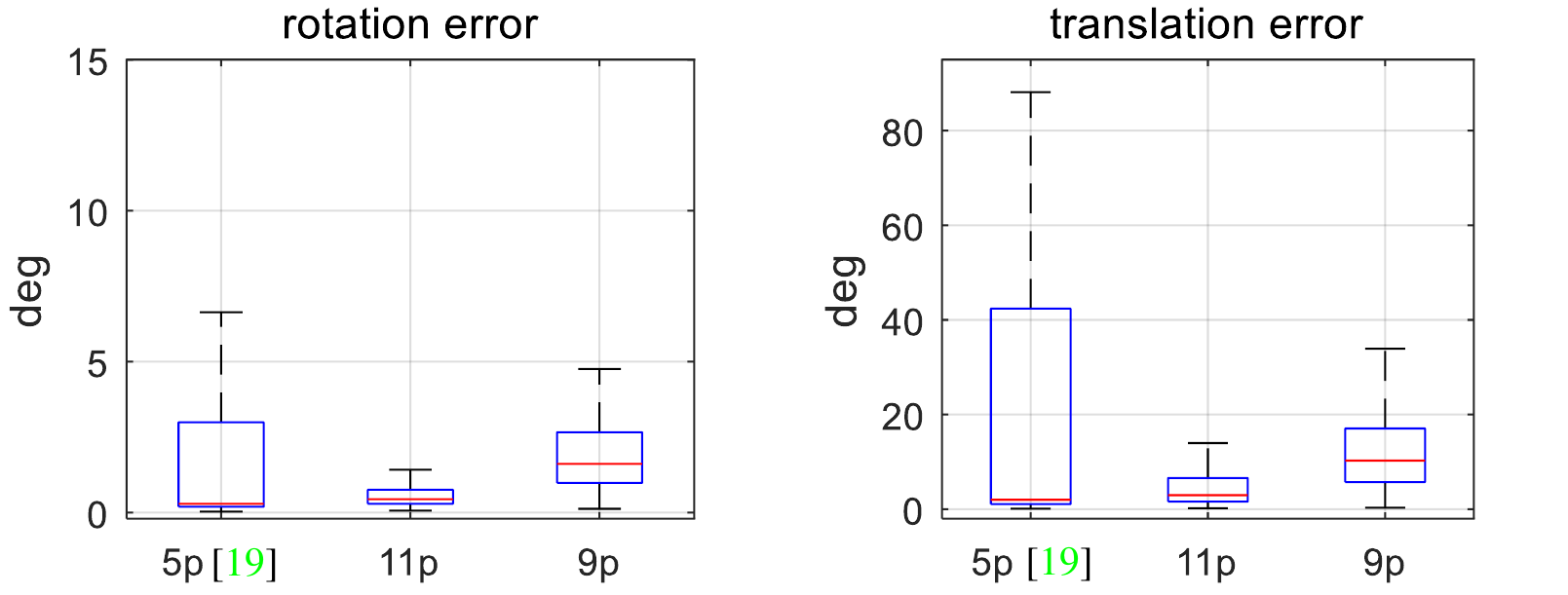}
			\caption{`as' sequence}
		\end{subfigure}
		\hspace{4mm}
		\begin{subfigure}[b]{0.99\linewidth}
			\includegraphics[width=\linewidth]{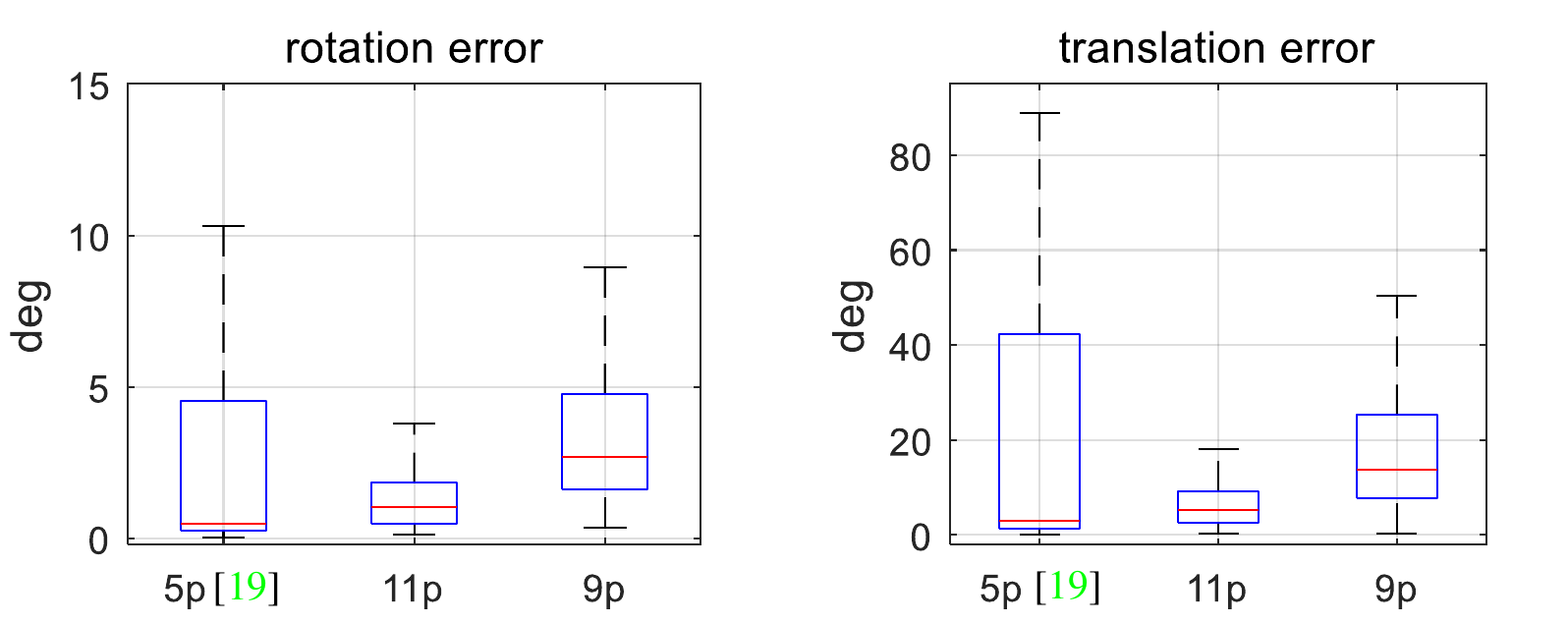}
			\caption{`af' sequence}		
		\end{subfigure} 
		\caption{Quantitative comparison for the PennCOSYVIO dataset (5p: the global shutter 5-point algorithm \cite{Nister:PAMI:2004}, 11p: the proposed uniform 11-point algorithm, 9p: the proposed uniform 9-point algorithm). }
		\label{fig:real_data_experimental_results_qunatitative}
	\end{figure}

	We evaluate the performance of the proposed algorithms using the public PennCOSYVIO dataset~\cite{Pfrommer:ICRA:2017}.
	The dataset was originally designed as a benchmark for performance evaluation of Visual-Inertial Odometry (VIO). 
	It provides image sequences, inertial measurements, and ground truth poses with a synchronized rolling shutter camera.
	The dataset consists mostly of forward motion and panning motion in the indoor and outdoor environments. 
	We perform experiments on forward motion because the panning motion is close to the pure rotation, which is a degenerate case of the relative pose estimation.
	The `as' (slow) and `af' (fast) sequences are used for our experiments.
	Each sequence is divided into 4 fragments.
	Table~\ref{table:real_data_details} describes the frame range, the number of image pairs, average linear and angular velocities used in the two sequences.	
	We extract image pairs every 5 frames within the frame range, and the interval of two frames is set to 30. 
	The number of images used in the experiment are 246 pairs for the `as' sequence and 212 pairs for `af' sequence.
	Figure~\ref{fig:real_data_sample_images} shows some sample images used for the experiment.		
	We use SIFT to extract and match features from the image pair.
	In order to reduce the IMU noise of the real dataset, we use an average of the inertial measurements (gravity and angular velocity) within 0.1$s$ near the image frame.

	Figure~\ref{fig:real_data_experimental_results_qunatitative} shows that the proposed 9- and 11-point algorithms outperform the 5-point algorithm in both `as' and `af' sequences.
	The rotation and translation estimates produced by the 5-point algorithm have larger variance than the estimates by the proposed algorithms.
	Especially, the variance of translation estimates is very large in both sequences. 
	The proposed 11-point algorithm produces more accurate estimates than the proposed 9-point algorithm because the hand-held dataset has severe gravity noises.
	Interestingly, the median value of the 5-point algorithm is very low.
	The reason is that this dataset has many image pairs with very small rolling shutter distortion as well.  	
	In such cases, the 5-point algorithm works well. 
	However, on average, the proposed 9- and 11-point algorithms produce more consistent and accurate estimates.

	\section{Conclusion}

	The rolling shutter distortion significantly affects the performance of geometric applications such as SfM and SLAM.
	In this paper, we exploited inertial measurements for practical use of the relative pose estimation of rolling shutter cameras.
	We proposed five different algorithms by applying the inertial measurements (gravity, angular velocity) to three rolling shutter camera models (linear, angular, and uniform).
	The synthetic and real data experiments show that the proposed methods produce more accurate relative pose than the conventional method.
	The proposed algorithms can be utilized to various geometric applications and can be extended to rolling shutter SfM and SLAM.
	
	{\small
		\bibliographystyle{ieee}
		\bibliography{rs}
	}

\end{document}